%% file: main.tex
\pdfoutput=1
\documentclass[11pt]{article}
\newcommand{\defcomment}[2]{#2}
\usepackage[]{acl}

\usepackage{times}
\usepackage{latexsym}

\usepackage[T1]{fontenc}
\usepackage[utf8]{inputenc}

\usepackage{microtype}

\usepackage[whole,ipaex]{bxcjkjatype} % for arXiv

\usepackage{graphicx}

\usepackage{url}
\usepackage{color}
\defcomment{
\newcommand{\draftonly}[1]{#1} 
}{%
\newcommand{\draftonly}[1]{}
}

\usepackage{gb4e} \noautomath
\usepackage{graphicx}
\usepackage{xspace}
\usepackage{enumitem}
\usepackage{here}
\renewcommand{\aa}{\={a}}
\newcommand{\ii}{ii}

\newcommand{\ou}{\={o}}
\usepackage{hyperref}
\newcommand{\ulabel}[1]{{\small\textsf{\uppercase{#1}}}} % label
\newcommand{\colora}[1]{{\color[rgb]{0, 0.6, 0} #1}} % generate
\newcommand{\colorb}[1]{{\color[rgb]{0, 0.6, 1} #1}} % annotate
\newcommand{\colorc}[1]{{\color[rgb]{0.9, 0, 0.6} #1}} % filter

\usepackage{hyperref}       % hyperlinks
\usepackage{booktabs}       % professional-quality tables
\usepackage{amsfonts}       % blackboard math symbols
\usepackage{nicefrac}       % compact symbols for 1/2, etc.
\usepackage{microtype}      % microtypography
\usepackage{xcolor}         % colors
\usepackage{pifont}
\usepackage{tikz}
\usepackage{soul}
\usepackage[whole]{bxcjkjatype} 
\usepackage{amsmath}
\usepackage{amsthm}
\usepackage{enumitem}

\usepackage{color_blind}    % see color_blind.sty

\definecolor{darkgreen}{HTML}{228B22}

\usepackage{tcolorbox}
\newtcolorbox{memobox}{
    colframe=cyan!20!white,
    colback =cyan!20!white,
    top=0mm, bottom=0mm, left=0mm, right=0mm,
    arc=0mm,
    fontupper=\color{blue!70!black},
    fonttitle=\bfseries\color{blue!70!black},
    title=Memo:
}

\newcommand{\kamiya}{\texttt{kamiya-codec}\xspace}
\newcommand{\total}{12,687\xspace} % total inflected forms
\newcommand{\pertotal}{118.6\xspace} % per word

\newcommand{\Riken}{$^{\heartsuit}$}
\newcommand{\Tohoku}{$^{\clubsuit}$}
\newcommand{\MBZUAI}{$^{\diamondsuit}$}

\newcommand{\dataname}{\textsc{J-UniMorph}\xspace}
\newcommand{\uni}{UniMorph\xspace}
\newcommand{\wik}{Wiktionary Edition\xspace}

\title{\dataname: Japanese Morphological Annotation \\ through the Universal Feature Schema}

\author{
    \textbf{Kosuke Matsuzaki}\Tohoku \quad
    \textbf{Masaya Taniguchi}\Riken  \quad
    \textbf{Kentaro Inui}\MBZUAI\Tohoku\Riken \quad
    \textbf{Keisuke Sakaguchi}\Tohoku\Riken \\
    \Tohoku Tohoku University \quad
    \Riken  RIKEN \quad
    \MBZUAI MBZUAI  \\
    \texttt{matsuzaki.kosuke.r7@dc.tohoku.ac.jp}\\
    \includegraphics[width=1em]{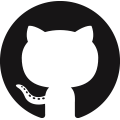}
\href{https://github.com/cl-tohoku/J-UniMorph}{github.com/cl-tohoku/J-UniMorph}
}

\begin{document}
\maketitle

\input{sections/0_abstract}

%\begin{exe}
%\ex 
%\gll Кот ест сметану\\
%cat.NOM eat.3.SG.PRS sour-cream.ACC\\
%\trans `The cat eats sour cream'
%\end{exe}

\input{sections/1_Introduction}

\input{sections/2_J-UniMorph}

\input{sections/3_Generate_forms}

\input{sections/4_Comparison}

\input{sections/5_Conclusion}

\input{sections/acknowledgments}

\bibliography{anthology,custom}
\bibliographystyle{acl_natbib}

\clearpage

\appendix
\input{sections/appendix}

\end{document}

%% file: sections/0_abstract.tex
\begin{abstract}
We introduce a Japanese Morphology dataset, \dataname, developed based on the UniMorph feature schema.
This dataset addresses the unique and rich verb forms characteristic of the language's agglutinative nature. 
\dataname distinguishes itself from the existing Japanese subset of UniMorph, which is automatically extracted from Wiktionary. 
On average, the Wiktionary Edition features around 12 inflected forms for each word and is primarily dominated by denominal verbs  (i.e., [noun] +\textit{suru} (do-\ulabel{PRS})). 
Morphologically, this form is equivalent to the verb \textit{suru} (do). 
In contrast, \dataname explores a much broader and more frequently used range of verb forms, offering 118 inflected forms for each word on average.
It includes honorifics, a range of politeness levels, and other linguistic nuances, emphasizing the distinctive characteristics of the Japanese language.
This paper presents detailed statistics and characteristics of \dataname, comparing it with the Wiktionary Edition. 
We release \dataname and its interactive visualizer publicly available, aiming to support cross-linguistic research and various applications.
\end{abstract}

%% file: sections/1_Introduction.tex
\section{Introduction}\label{sec:intro}

\input{figures/generate_flow}

Universal Morphology (\uni) is a collaborative project that delivers a wide-ranging collection of standardized morphological features for over 170 languages in the world~\citep{schema:16,UniMorph3.0:20}.
\uni feature schema comprises over 212 feature labels across 23 dimensions of meaning labels, such as tense, aspect, and mood. 
More concretely, \uni dataset consists of a lemma coupled with a set of morphological features that correspond to a specific inflected form, as illustrated by the following example:
\vspace{-2mm}
\begin{table}[H]
    \centering
    \setlength{\tabcolsep}{5pt}
    \begin{tabular}{lll}
        走る/\textit{hashi-ru} & 走った/\textit{hashi-tta} & \ulabel{V;PST;IPFV}
    \end{tabular}
\end{table}%
\vspace{-3.5mm}

\noindent
where the original form (lemma)
``\textit{hashi-ru}'' (走る, run-\ulabel{prs}) is inflected to ``\textit{hashi-tta}'' (走った, run-\ulabel{pst})
to indicate the past tense (\ulabel{PST}) and imperfective aspect (\ulabel{IPFV}) as morphological features.

The challenge of morphological (re)inflection, which started with the SIGMORPHON 2016 Shared Task~\cite{sigmorphon:16}, involves generating an inflected form from a given lemma and its corresponding morphological feature.
This effort has continued over years, covering multiple shared tasks~\cite{sigmorphon:17,sigmorphon:18,sigmorphon:19,sigmorphon:20,sigmorphon:21,sigmorphon:22,sigmorphon:23}.

The SIGMORPHON–UniMorph 2023 Shared Task 0~\cite{sigmorphon:23} released a Japanese Morphology dataset,\footnote{\url{https://github.com/sigmorphon/2023InflectionST/}} which was automatically extracted from Wiktionary. 
This Wiktionary Edition, on average, highlights 12 inflected forms for each word. 
It mainly consists of denominal verbs, which are formed by combining a noun with ``\textit{suru}'' (do-\ulabel{prs}), and their inflection patterns are morphologically equivalent to those of the verb ``\textit{suru}.''

We introduce \dataname, which emphasizes a wider variety of verb forms, with an average of 118 inflected forms per word.
It includes honorifics, varying levels of politeness, and imperatives with fine-grained distinctions, showcasing the distinctive features of the Japanese language.

This paper begins with an overview of the UniMorph Schema, detailing the characteristics of each dimension and the criteria used for labeling \dataname (\S\ref{sec:feature}). 
We then explain the data creation process for \dataname (\S\ref{sec:generate}).
As illustrated in Figure~\ref{fig:generate_flow}, this process includes three main steps: (1) generating inflected forms (\colora{Generation}), (2) assigning UniMorph labels (\colorb{Annotation}), and (3) removing incorrect or infrequent forms based on frequency  (\colorc{Filtering}). 
Finally, a comparative analysis (\S\ref{sec:compare}) between \dataname and the Wiktionary Edition shows that \dataname includes more commonly used verbs and a wider variety of inflected forms than the \wik, with a slightly larger size (\total vs. 12,000).

We release \dataname and its interactive visualizer publicly available, aiming to provide a useful resource for cross-linguistic studies and a range of applications.

%% file: figures/generate_flow.tex
\begin{figure*}[t]
    \centering
    \includegraphics[width=16cm]{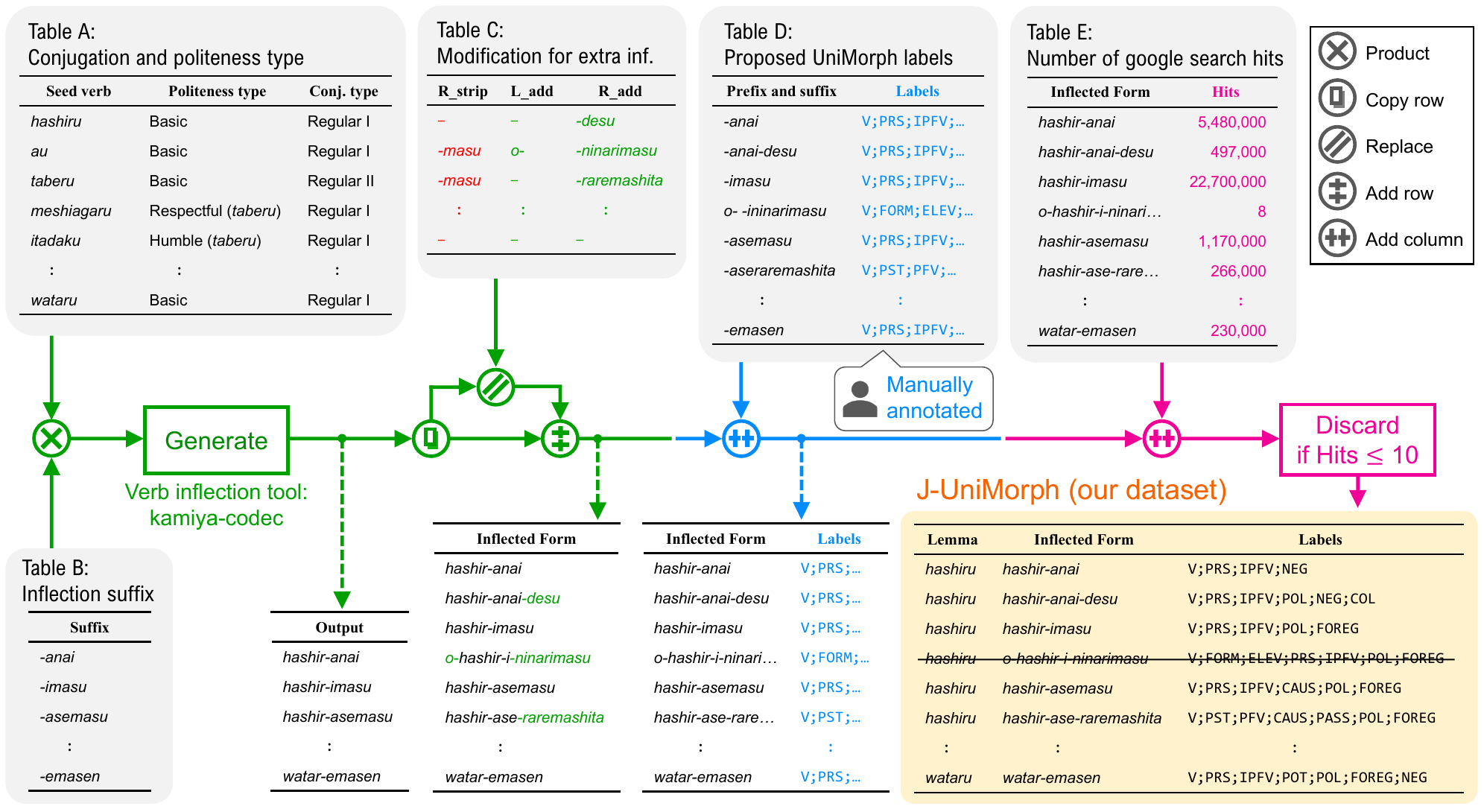}
    \caption{
    Overview of the \dataname creation process:
    First, we \colora{generate} inflected forms from seed verbs (Table A, detailed in \S\ref{subsec:criteria}) and inflection suffix (Table B, detailed in \S\ref{subsec:generate}) using the verb inflection tool, \kamiya. 
    This is followed by modifying and adding inflected forms that the tool does not cover
    (Table C, detailed in \S\ref{subsec:generate}).
    Second, Japanese native speakers \colorb{annotate} \uni labels to each form (Table D, detailed in \S\ref{sec:feature}). 
    Finally, we apply a frequency \colorc{filter} to discard infrequent inflected forms (Table E, detailed in \S\ref{subsec:filtering}).
    }
    \label{fig:generate_flow}
\end{figure*}

%% file: sections/2_J-UniMorph.tex
\section{Features Schema in \dataname}\label{sec:feature}

Japanese, an agglutinative language, allows for the expression of various meanings through altering verb endings and adding affixes.
Verbs in Japanese are broadly categorized into three conjugation types: Regular I verbs, Regular II verbs, and Irregular verbs~\cite{KamiyaVerb:01}. 
Among these, the Irregular verbs include only ``\textit{kuru}'' (来る, come-\ulabel{prs}) and ``\textit{suru}'' (する, do-\ulabel{prs}).\footnote{In Japanese, \textit{denominal verbs} are formed by combining a noun with the light verb ``\textit{suru}.'' 
For example, ``\textit{benkyo}'' (勉強, study-\ulabel{n}) becomes ``\textit{benkyo-suru}'' (勉強する, study-\ulabel{v;prs}).
These verbs share the same inflection pattern as ``\textit{suru}'' (do-\ulabel{v;prs}).
Given their identical inflection pattern, we have excluded denominal verbs from the \dataname.}
Table~\ref{tab:regular_verbs_ex} provides examples of Regular I and II verbs.

\input{tables/Regular_I,II_examples}

The authors, who are all native Japanese speakers with Linguistics backgrounds, have carefully and thoroughly discussed to determine the alignment between the inflection patterns and their \uni feature labels.\footnote{The ``label'' is also referred to as ``tag'' recently~\cite{UniMorph3.0:20,UniMorph4.0:22}.}
In this section, we review the common Japanese inflections such as
politeness (\S\ref{subsec:2politeness}),
mood including imperatives (\S\ref{subsec:2mood}),
tense and aspect (\S\ref{subsec:2tense}),
negation (\S\ref{subsec:2negation}),
passive (\S\ref{subsec:2passive}),
and causative (\S\ref{subsec:2causative}), and the criteria for labeling \dataname.
We note that some inflected forms share the same spelling but have ambiguous or multiple meanings, and we annotate these as distinct entries in \dataname for clarity.

\subsection{Politeness}
\label{subsec:2politeness}

Honorific speech (\textit{Keigo}, 敬語), which conveys politeness in Japanese, is primarily classified into three types:
polite form (\textit{Teineigo}, 丁寧語), respectful form (\textit{Sonkeigo}, 尊敬語), and humble form (\textit{Kenj\ou go}, 謙譲語).
We explain the characteristics, forms of expression, usage, and applicable labels in the following.

\paragraph{Polite form (\textit{Teineigo}, 丁寧語)}

Polite form is a form that conveys respect to the reader or listener, and it uses the ``\textit{-desu/masu}'' (-です/ます) form. 
The level of politeness can be further heightened when used in inflection with respectful or humble form~\cite{keigo:88}. 
The \uni Schema includes the label \ulabel{pol} (Polite), so we attach this label to these form. Additionally, the \uni Schema provides the label \ulabel{foreg} (Formal register) for the Japanese ``\textit{mas(u)}-style''~\cite{schema:16}; therefore we have also assigned \ulabel{foreg} to the ``\textit{-masu}'' form.

\paragraph{Respectful form (\textit{Sonkeigo}, 尊敬語)}

The respectful form of expression elevates the person who should be respected, and is typically used for superiors and customers.
This is not employed for individuals within the same group or for one's own actions.
Most verbs generally take the form of ``\textit{-re-ru/rare-ru}'' (-れる/られる), and ``\textit{o---ninaru}'' (お---になる), where the verb's inflection occurs between the ``\textit{o}'' and ``\textit{ninaru}.''
Some verbs also take lexical honorifics, where the word itself changes to express respect, such as changing ``\textit{iku}'' (行く, go-\ulabel{prs}\footnote{In the main text, only the relevant label set is presented for brevity.}) to ``\textit{irassharu}'' (いらっしゃる, go-\ulabel{prs;elev}).

Although these lexical honorifics go beyond the scope of standard ``inflection,'' we have chosen to include some of them for practical reasons, particularly because they are commonly used in place of basic verbs when expressing respect.

The~``\textit{o---ninaru}'' (お---になる) form is commonly used for verbs that do not have any lexical honorific.
Both the lexical honorific and the ``\textit{o---ninaru}'' form are labeled with \ulabel{form+elev} (Formal, Referent Elevating), following the \uni Schema~\cite{schema:16}.
The~``\textit{-re-ru/rare-ru}'' form is assigned only \ulabel{elev} (Referent Elevating) without \ulabel{form} (Formal). 
This choice is based on the consideration that this form conveys a lower level of respect compared to the ``\textit{o---ninaru}'' (お---になる) and the lexical honorific, despite slightly deviating from the \uni Schema definition~\cite{schema:16}.
The following examples illustrate the verb ``\textit{iku}'' (行く, go-\ulabel{prs}) with a lexical honorific and ``\textit{au}'' (会う, meet-\ulabel{prs}) without a lexical honorific.

\begin{table}[H]
    \setlength{\tabcolsep}{3pt}
    \begin{tabular}{ll}
        行く/\textit{iku} & 行く/\textit{iku}\\
        行かれる/\textit{ika-reru} & いらっしゃる/\textit{irassharu}\\
        \ulabel{v;prs;ipfv;\textbf{elev}} & \ulabel{v;\textbf{form};\textbf{elev};prs;ipfv}\\
        & \\
        会う/\textit{au} & 会う/\textit{au}\\
        会われる/\textit{awa-reru} & お会いになる/\textit{o-ai-ninaru}\\
        \ulabel{v;prs;ipfv;\textbf{elev}} & \ulabel{v;\textbf{form};\textbf{elev};prs;ipfv}\\
    \end{tabular}
\end{table}

\input{tables/imperative_type}

\paragraph{Humble form (\textit{Kenj\ou go}, 謙譲語)}

The humble form conveys respect by lowering oneself or one's group in comparison to the person deserving respect.
In business contexts, it is used even when referring to the actions of one's own company's superiors, especially when addressing customers. 
Most verbs mainly take the form of ``\textit{o---suru}'' (お---する), where the verb's inflection occurs between the ``\textit{o}'' and ``\textit{suru}.'' Some verbs also take lexical honorifics.
These are labeled as \ulabel{form+humb} (Formal, Speaker Humbling), following the \uni Schema~\cite{schema:16}.
The examples below demonstrate the use of the verb ``\textit{iku}'' (行く, go-\ulabel{prs}) with the lexical honorific and ``\textit{kaku}'' (書く, write-\ulabel{prs}) without a lexical honorific.

\begin{table}[H]
    \begin{tabular}{l}
        行く/\textit{iku}\\
        伺う/\textit{ukagau}\\
        \ulabel{v;\textbf{form};\textbf{humb};prs;ipfv}\\
        \\
        書く/\textit{kaku}\\
        お書きする/\textit{o-kaki-suru}\\
        \ulabel{v;\textbf{form};\textbf{humb};prs;ipfv}\\
    \end{tabular}
\end{table}
\vspace{-2mm}

The complexity of Japanese honorifics and their inflection patterns is further complicated by lexical honorifics corresponding to multiple basic forms, and vice versa.
For instance, the humble verb ``\textit{ukagau}'' (伺う) corresponds to three basic verbs: ``\textit{kuru}'' (来る, come), ``\textit{iku}'' (行く, go), and ``\textit{kiku}'' (聞く, ask/listen).
On the other hand, the basic verb ``\textit{iku}'' (行く, go) is associated with three humble verbs: ``\textit{mairu}'' (まいる), ``\textit{ukagau}'' (伺う), and ``\textit{agaru}'' (上がる).
In Appendix \ref{app:keigo}, we provide the correspondence between the basic forms and lexical honorifics adopted in \dataname.

\subsection{Mood}
\label{subsec:2mood}

In terms of expressing mood, we deal with the following five categories: Imperatives, Intentive, Optative, Potential, and Permissive.

\paragraph{Imperatives}

Japanese has a variety of imperative expressions, as shown in Table~\ref{tab:types_of_imperative}.
This table compiles the inflection and label correspondence of the verb ``\textit{tabe-ru}'' (食べる, eat-\ulabel{PRS}) as an example, organizing them into four groups based on the similarity of their label sets. 
Each group's inflected forms are roughly sorted by  the strength of degree of command, from strong to weak.
All forms in Table \ref{tab:types_of_imperative} are labeled \ulabel{imp} (Imperative).

In Table \ref{tab:types_of_imperative}, the term ``\textit{tabe-ro}'' (食べろ, Do eat!), representing the most forceful command, is annotated with \ulabel{oblig} (Obligative) due to its compelling nature.
This expression is rarely used in everyday conversations as it comes across as overly authoritative.
For colloquial forms like ``\textit{tabe-na}'' (食べな, Eat.) in the second row, commonly used in informal speech, \ulabel{col} (Colloquial) is assigned.
For forms that include polite expressions such as ``\textit{-nasai}'' and ``\textit{-kudasai},'' \ulabel{pol} (Polite) is assigned.

The bottom two groups of Table~\ref{tab:types_of_imperative} show imperative inflection patterns and their corresponding labels for lexical honorifics  ``\textit{meshiaga-ru}'' (召し上がる, eat-\ulabel{prs;elev}), which is one of the respectful forms
of the basic verb ``\textit{tabe-ru}'' (食べる, eat-\ulabel{prs}).
For these instances, we also assign \ulabel{form+elev} labels (\S\ref{subsec:2politeness}).

\paragraph{Intentive}

Japanese intentive forms such as ``\textit{-y\ou}'' (-よう), ``\textit{-\ou}'' (-おう), and ``\textit{-mash\ou}'' (-ましょう) are marked with \ulabel{inten} (Intentive).
Since ``\textit{-mash\ou}'' is one of the inflections of the polite form ``\textit{-masu}'' (-ます), it is additionally annotated with \ulabel{pol+foreg} (Polite, Formal register) (\S\ref{subsec:2politeness}).
Below are examples of intentive expressions, where ``\textit{tabe-y\ou}'' and ``\textit{tabe-mash\ou}'' are inflected forms of the verb ``\textit{tabe-ru}'' (食べる, eat-\ulabel{PRS}).

\begin{table}[H]
    \begin{tabular}{l}
        ピザを食べよう。\\
        \textit{Piza-o tabe-\textbf{y\ou}.}\\
        Let's eat pizza.\\
    \end{tabular}
\end{table}
\vspace{-5mm}
\begin{table}[H]
    \begin{tabular}{l}
        ピザを食べましょう。\\
        \textit{Piza-o tabe-\textbf{mash\ou}.}\\
        Let's eat pizza. (Polite)
    \end{tabular}
\end{table}

\vspace{-3mm}
\paragraph{Optative}

In Japanese, expressions of wishes include ``\textit{-tai}'' (たい) to express subjective desires and ``\textit{-tagaru}'' (たがる) for objective observations.
We distinguish these two optative expressions with the label \ulabel{opt} (Optative-Desiderative), associated with person specification (\ulabel{1}: first person, \ulabel{3}: third person).
Below are examples with the verb ``\textit{hashir-u}'' (走る, run).

\begin{table}[H]
\setlength{\tabcolsep}{0pt}
    \begin{tabular}{l}
        走る/\textit{hashir-u} \\
        走りたい/\textit{hashir-i-tai} \\
        \ulabel{v;prs;ipfv;opt;1}\\
        e.g., I want to run. (\textit{Watashi-wa hashir-i-\textbf{tai}})\\
        
    \end{tabular}
\end{table}
\vspace{-5mm}
\begin{table}[H]
    \setlength{\tabcolsep}{0pt}
    \begin{tabular}{l}
        走る/\textit{hashir-u} \\
        走りたがる/\textit{hashir-i-tagaru} \\
        \ulabel{v;prs;ipfv;opt;3}\\
        e.g., He wants to run. (\textit{Kare-wa hashir-i-\textbf{tagaru}})
    \end{tabular}
\end{table}

\paragraph{Potential}

We assign the label \ulabel{pot} (Potential) to expressions that indicate possibility. 
For Regular I verbs, the suffix ``\textit{-eru}'' is attached, while Regular II verbs take ``\textit{-reru/rareru},''
which is identical to the respectful form (\S\ref{subsec:2politeness}). 
In \dataname, we include these forms as separate entries.
Below are examples, with ``\textit{kaku}'' (書く, write-\ulabel{prs}) being a Regular I verb and ``\textit{miru}'' (見る, look-\ulabel{prs}) a Regular II verb.

\begin{table}[H]
    \begin{tabular}{lll}
        書く/\textit{kak-u} & 見る/\textit{mi-ru}\\
        書ける/\textit{kak-eru} & 見られる/\textit{mi-rareru}\\
        \ulabel{v;prs;ipfv;\textbf{pot}} & \ulabel{v;prs;ipfv;\textbf{pot}}
    \end{tabular}
\end{table}

\paragraph{Permissive}

The expression ``\textit{-(sa)se-te-itadaku}'' (-させていただく) is used to politely request permission, demonstrating humility.\footnote{While originally meant for contexts where a specific approver for a particular action could be anticipated, it has now changed to express humility even when the approver may not be evident~\cite{Nitta7:09}.}
We assigned this form with \ulabel{form+humb} (Formal, Speaker Humbling) and \ulabel{perm} (Permissive).
The following examples demonstrate annotated suffixes for ``\textit{-(sa)se-te-itadaki-masu}'' with \ulabel{v;\textbf{form};\textbf{humb};prs;ipfv;\textbf{pol};\textbf{foreg};\textbf{perm}}.

\begin{itemize}
    \item[(a)] 私から答えさせていただきます。\\
    \textit{Watashi-kara kotae-\textbf{sase-te-itadaki-masu}.}\\
    (If I am allowed,) I will answer (the question).\footnote{Brackets indicate implied meaning not explicitly stated in Japanese.}
    \item[(b)] [店先の貼り紙で] 本日は休ませていただきます．\\
    \textit{Honjitsu-wa yasuma-\textbf{se-te-itadaki-masu}.}\\
    \textrm{[}Notice at the store front] (Our store) will be closed today. (No specific permission is required)
\end{itemize}

\subsection{Tense and Aspect}
\label{subsec:2tense}

There are two forms to express tense or aspect: \textit{ta}-form and \textit{ru}-form in Japanese.
The ``\textit{ta}'' and ``\textit{ru}'' respectively represent verb endings such as ``\textit{tabe-ta}'' (食べた, eat-\ulabel{pst}) or ``\textit{tabe-ru}'' (食べる, eat-\ulabel{prs}). 
From a tense perspective, these forms represent the contrast between ``past'' and ``non-past,'' while from an aspect perspective, they represent the contrast between ``perfective'' and ``imperfective''~\cite{tense_aspect:89}.

Japanese does not have a distinct form to explicitly distinguish between present and future.
Future tense is expressed by adverbial elements such as ``next week'' or ``tomorrow,'' so we do not assign the label \ulabel{fut} (Future) to the \textit{ru}-form.

Based on the above considerations, the \textit{ta}-form is assigned the label \ulabel{pst+pfv} (Past, Perfective), while the \textit{ru}-form is assigned the label \ulabel{prs+ipfv} (Present, Imperfective).
The following are examples of the verb ``\textit{hashi-ru}'' (走る, run-\ulabel{prs}).\footnote{As in this example, the \textit{ta}-form does not necessarily involve simply replacing ``\textit{ru}'' with ``\textit{ta}'' from the base form.}

\begin{table}[H]
    \begin{tabular}{ll}
        走る/\textit{hashi-ru} & 走る/\textit{hashi-ru}\\
        走る/\textit{hashi-ru} & 走った/\textit{hashi-tta}\\
        \ulabel{v;prs;ipfv} & \ulabel{v;pst;pfv}
    \end{tabular}
\end{table}%

Note that the \textit{ru}- and \textit{ta}-form have various meanings by being accompanied by peripheral words such as adverbs and interjections.
The examples about special usage of the \textit{ru}-form are
property: 日本人は米を\textbf{食べる}。(Japanese people eat rice.), 
and command: さっさと\textbf{歩く}！ (Walk quickly!).
The examples about special usage of the \textit{ta}-form are
discovery: [鍵を探していて] あっ、ここに\textbf{あった}。(Oh, here's the key.),
and recall: あっ、今日は会議\textbf{だった}。(Oh, I have a meeting today.)~\cite{Nitta3:07}.
Since the meaning of these cases relies on peripheral words, not on the inflected form itself, we exclude these instances from the \dataname.

Prospective forms such as ``\textit{-dar\ou}'' (-だろう) and ``\textit{-desh\ou}'' (-でしょう) are marked with \ulabel{prosp} (Prospective). As ``\textit{-desh\ou}'' is one of the inflections of the polite form ``\textit{-desu},'' it is also annotated with \ulabel{pol} (Polite). An example of the usage of ``\textit{-desh\ou}'' is presented below.

\begin{itemize}
    \item[] 明日は晴れるでしょう。\\
    \textit{Ashita-wa hare-ru-\textbf{desh\ou}.}\\
    It will be sunny tomorrow.
\end{itemize}

\subsection{Negation}
\label{subsec:2negation}

Negation in Japanese is primarily expressed  through the suffixes ``\textit{-nai}'' (-ない) or ``\textit{-masen}'' (-ません), and in \dataname, the label \ulabel{neg} (Negative) is assigned to indicate negation. 
Since ``\textit{-masen}'' is an inflection of the polite form ``\textit{-masu},''
we assign the label \ulabel{pol+foreg+neg} (Polite, Formal register, Negative) to it. 
Another polite negation form, ``\textit{-nai-desu}'', is commonly used in colloquial speech, and thus, the label \ulabel{pol+neg+col} (Negative, Colloquial) is applied to it. 

Importantly, neither ``\textit{-nai}'' (\ulabel{neg}) nor ``\textit{-desu}'' (\ulabel{pol}) alone conveys a colloquial tone; however, \ulabel{col} becomes apparent when they are combined, highlighting the non-monotonic compositional nature of verb inflection in Japanese. 
Below are examples of ``\textit{mi-ru}'' (見る, look-PRS).

\vspace{-1mm}
\begin{table}[H]
    \begin{tabular}{ll}
        見る/\textit{mi-ru} & 見る/\textit{mi-ru}\\
        見ない/\textit{mi-nai} & 見ないです/\textit{mi-nai-desu}\\
        \ulabel{v;prs;ipfv;\textbf{neg}} & \ulabel{v;prs;ipfv;\textbf{pol};\textbf{neg};\textbf{col}}
    \end{tabular}
\end{table}
\vspace{-5mm}
\begin{table}[H]
    \begin{tabular}{ll}
        見る/\textit{mi-ru} & \\
        \multicolumn{2}{l}{見ません/\textit{mi-masen}}\\
        \multicolumn{2}{l}{\ulabel{v;prs;ipfv;\textbf{pol};\textbf{foreg};\textbf{neg}}}
    \end{tabular}
\end{table}

\subsection{Passive}
\label{subsec:2passive}

The passive voice (\ulabel{pass}) in Japanese employs the suffix ``\textit{-re-ru/rare-ru}'' (-れる/られる), which shares the same form as the respectful form (\S\ref{subsec:2politeness} and also potential form (\S\ref{subsec:2mood}).
In \dataname, we categorize these forms as distinct entries for clarity.
The past expression is created by replacing the last ``\textit{ru}'' with ``\textit{ta}'' (\S\ref{subsec:2tense}), resulting in ``-\textit{re-ta/rare-ta}'' (-れた/られた).
An example of the use of past and passive expression is provided below. 

\vspace{-1mm}
\begin{table}[H]
    \begin{tabular}{l}
    私のテスト用紙を彼に見られた。\\
    \textit{Watashi-no tesuto y\ou shi-o kare-ni mi-\textbf{rare-ta}.}\\
    My test paper was seen by him.
    \end{tabular}
\end{table}

\subsection{Causative}
\label{subsec:2causative}

In English, causatives are typically expressed using verbs like ``have'' or ``make.'' However, in Japanese, this can be achieved using suffixes, specifically the ``-\textit{se-ru/sase-ru}'' (-せる/させる) form, which is annotated with \ulabel{caus} (Causative).\footnote{We explain \textit{lexical} causative verbs in \S\ref{subsec:2future}.}
The past expression is created by replacing the suffix ``\textit{ru}'' with ``\textit{ta}'' (\S\ref{subsec:2tense}), resulting in ``-\textit{se-ta/sase-ta}'' (-せた/させた). Below is an example of the past expression.
\vspace{-4mm}
\begin{table}[H]
    \begin{tabular}{l}
    私はその映画を彼に見させた。\\
    \textit{Watashi-wa sono eiga-o kare-ni mi-\textbf{sase-ta}.}\\
    I made him watch the movie.
    \end{tabular}
\end{table}
\vspace{-2mm}

We also deal with the following forms: causative involving passive, and contraction of causative.

\paragraph{Causative and Passive}

It is possible to introduce passivity into the causative construction. In such cases, the ``-\textit{se-rare-ru/sase-rare-ru}'' (-せられる/させられる) form is employed, and the labels are annotated with \ulabel{caus+pass} (Causative, Passive). The past expression is created by replacing the last ``\textit{ru}'' with ``\textit{ta}'' (\S\ref{subsec:2tense}), resulting in ``-\textit{se-rare-ta/sase-rare-ta}'' (-せられた/させられた). Below is an example of the past expression.

\begin{table}[H]
\setlength{\tabcolsep}{0pt}
    \begin{tabular}{l}
    私はその映画を彼に見させられた。\\
    \textit{Watashi-wa sono eiga-o kare-ni mi-\textbf{sase-rare-ta}.}\\
    I was made to watch the movie by him.\\
    $\approx$ He made me watch the movie.
    \end{tabular}
\end{table}

\paragraph{Contraction of Causative}

\input{tables/ex_causative}

The contracted form ``\textit{-su/sasu}'' (-す/さす) is frequently used for causative verbs. 
In Regular I Verbs, similarly, the contracted form ``\textit{-sare-ru}'' (-される) is commonly employed for passive-causative expression~\cite{Nitta2:09}. 
Examples of each are presented in Table~\ref{tab:causative_ex} and Table~\ref{tab:passive_causative_ex}.

These shortening forms like ``\textit{-su/sasu}'' or ``\textit{-sare-ru}'' are assigned the same labels as ``\textit{-se-ru/sase-ru}'' (\ulabel{caus}) or ``\textit{-se-rare-ru/sase-rare-ru}'' (\ulabel{caus+pass}), because they do not lead to any change in meaning, such as a decrease in respect.
Below are examples of causative of ``\textit{tabe-ru}'' (食べる, eat-\ulabel{prs}).

\begin{table}[H]
    \setlength{\tabcolsep}{4pt}
    \begin{tabular}{ll}
        食べる/\textit{tabe-ru} & 食べる/\textit{tabe-ru}\\
        食べさせる/\textit{tabe-sase-ru} & 食べさす/\textit{tabe-sasu}\\
        \ulabel{V;PRS;IPFV;\textbf{CAUS}} & \ulabel{V;PRS;IPFV;\textbf{CAUS}}
    \end{tabular}
\end{table}

%% file: tables/Regular_I,II_examples.tex
\begin{table}[t]
    \centering
    \footnotesize
    \begin{tabular}{ll}
        \toprule
        \multicolumn{2}{l}{Regular I verbs (I型動詞，五段活用動詞)} \\
        & \textit{a-u} (会う, meet), \textit{ik-u} (行く, go), \textit{kak-u} (書く, write),\\
        & \textit{kik-u} (聞く, listen), \textit{hashir-u} (走る, run)  \\
        \midrule
        \multicolumn{2}{l}{Regular II verbs (II型動詞，一段活用動詞)} \\
        & \textit{ki-ru} (着る, wear/put on), \textit{kotae-ru} (答える, answer),\\
        & \textit{tabe-ru} (食べる, eat), \textit{mi-ru} (見る, see/watch) \\
        \bottomrule
    \end{tabular}
    \caption{Examples of Regular I and II Verbs}
    \label{tab:regular_verbs_ex}
\end{table}

%% file: tables/imperative_type.tex
\begin{table*}[t]
  \centering
  \footnotesize
  \setlength{\tabcolsep}{5pt}
  \begin{tabular}{llll}
    \toprule
    Inflected form & Romanization & Labels \\
    \midrule
    食べろ & \textit{tabe-ro} & \ulabel{v;imp;oblig} \\
    食べな & \textit{tabe-na} & \ulabel{v;imp;oblig;col} \\
    食べなさい & \textit{tabe-nasai} & \ulabel{v;imp;oblig;pol} \\
    食べて & \textit{tabe-te} & \ulabel{v;imp;col} \\
    食べてください & \textit{tabe-te-kudasai} & \ulabel{v;imp;pol} \\
    お食べください & \textit{o-tabe-kudasai} & \ulabel{v;form;imp;pol} \\
    \midrule
    食べるな & \textit{tabe-ru-na} & \ulabel{v;imp;oblig;neg} \\
    食べないで & \textit{tabe-nai-de} & \ulabel{v;imp;neg;col} \\
    食べないでください & \textit{tabe-nai-de-kudasai} & \ulabel{v;imp;pol;neg} \\
    お食べにならないでください & \textit{o-tabe-ni-naranai-de-kudasai} & \ulabel{v;form;imp;pol;neg} \\
    \midrule
    召し上がれ & \textit{meshiaga-re} & \ulabel{v;form;elev;imp;oblig} \\
    召し上がりな & \textit{meshiaga-ri-na} & \ulabel{v;form;elev;imp;oblig;col} \\
    召し上がりなさい & \textit{meshiaga-ri-nasai} & \ulabel{v;form;elev;imp;oblig;pol} \\
    召し上がって & \textit{meshiaga-tte} & \ulabel{v;form;elev;imp;col} \\
    召し上がってください & \textit{meshiaga-tte-kudasai} & \ulabel{v;form;elev;imp;pol} \\
    お召し上がりください & \textit{o-meshiaga-ri-kudasai} & \ulabel{v;form;elev;imp;pol;col} \\
    \midrule
    召し上がるな & \textit{meshiaga-ru-na} & \ulabel{v;form;elev;imp;oblig;neg} \\
    召し上がらないで & \textit{meshiaga-ra-nai-de} & \ulabel{v;form;elev;imp;neg;col} \\
    召し上がらないでください & \textit{meshiaga-ra-nai-de-kudasai} & \ulabel{v;form;elev;imp;pol;neg} \\
    お召し上がりにならないでください & \textit{o-meshiaga-ri-ni-naranai-de-kudasai} & \ulabel{v;form;elev;imp;pol;neg;col} \\
    \bottomrule
  \end{tabular}
  \caption{Correspondence between the imperative form and labels, using the verb ``\textit{taberu}'' (食べる, eat).}
  \label{tab:types_of_imperative}
\end{table*}

%% file: tables/ex_causative.tex
\begin{table}[t]
    \centering
    \footnotesize
    \begin{tabular}{llll}
        \toprule
        Conj. type & Base & Ordinary & Contraction \\
        \midrule
        Reg. I & 書く & 書かせる & 書かす\\
        & \textit{kak-u} & \textit{kak-ase-ru} & \textit{kak-as-u}\\
        Reg. II & 見る & 見させる & 見さす\\
        & \textit{mi-ru} & \textit{mi-sase-ru} & \textit{mi-sas-u}\\
        Irreg. & 来る & 来させる & 来さす\\
        & \textit{ku-ru} & \textit{ko-sase-ru} & \textit{ko-sas-u}\\
        Irreg. & する & させる & さす\\
        & \textit{su-ru} & \textit{s-ase-ru} & \textit{sas-u}\\
        \bottomrule
    \end{tabular}
    \caption{Examples of Causative contraction forms. We also handle these contraction forms.}
    \label{tab:causative_ex}
\end{table}

\begin{table}[t]
    \centering
    \footnotesize
    \setlength{\tabcolsep}{5pt}
    \begin{tabular}{llll}
        \toprule
        Conj. type & Base & Ordinary & Contraction \\
        \midrule
        Reg. I & 書く & 書かせられる & 書かされる\\
        & \textit{kak-u} & \textit{kak-ase-rare-ru} & \textit{kak-as-are-ru}\\
        Reg. II & 見る & 見させられる & *見さされる\\
        & \textit{mi-ru} & \textit{mi-sase-rare-ru} & *\textit{mi-sas-are-ru}\\
        Irreg. & 来る & 来させられる & *来さされる\\
        & \textit{ku-ru} & \textit{ko-sase-rare-ru} & *\textit{ko-sas-are-ru}\\
        Irreg. & する & させられる & *さされる\\
        & \textit{su-ru} & \textit{s-ase-rare-ru} & *\textit{sas-are-ru}\\
        \bottomrule
    \end{tabular}
    \caption{Examples of Passive-Causative contraction forms. We do not handle incorrect usages, which have the asterisk (*).}
    \label{tab:passive_causative_ex}
\end{table}

%% file: sections/3_Generate_forms.tex
\section{How to Generate Inflected Forms}
\label{sec:generate}

The previous section outlined how we matched inflected forms with their \uni labels. 
In this section, we will walk through our process for generating all the inflected forms and how we filter out the less common forms, yielding a total of \total in \dataname.

\subsection{Seed Verb Selection Process}
\label{subsec:criteria}

The selection of seed verbs (Table A in Figure~\ref{fig:generate_flow}) comprised two categories: (a) 107 basic verbs frequently encountered at the N5 (most basic) level of the Japanese Language Proficiency Test (JLPT), and (b) 40 lexical honorifics,\footnote{Lexical honorifics are matched with the corresponding 107 basic verbs.} divided into 19 respectful and 21 humble forms, as cited in~\citet{keigo:88}.
The conjugation types of each verb and their detailed statistics are provided in Appendix~\ref{app:verb_count}.

\subsection{Generate Inflected Forms}
\label{subsec:generate}

First, we made a list of inflected forms to be registered in \dataname (Table B in Figure~\ref{fig:generate_flow}).\footnote{In most cases, the inflected forms correspond to the ``\textit{bunsetsu},'' Japanese grammatical unit which is roughly equivalent to a (verb) phrase in English. However, they can sometimes extend beyond a single ``\textit{bunsetsu},'' especially when multiple suffixes are combined (cf. \citet{MightyMorph:22}).}
Inflected forms were carefully selected by four native speakers of Japanese (the authors), who referred to several books on Japanese grammar~\cite{Nitta3:07,Nitta2:09,Nitta7:09,keigo:88,PassCaus:11} and a book designed for Japanese language learners~\cite{KamiyaVerb:01}.

Next, we used \kamiya,\footnote{\url{https://github.com/fasiha/kamiya-codec}} a verb inflection tool, to generate each inflected form based on patterns derived from~\citet{KamiyaVerb:01}.
This tool produces inflected forms by taking the seed verb (lemma) and the arguments for its inflections.\footnote{One exception is the negation of ``\textit{ar-u}'' (ある, be), which is expressed as ``\textit{nai}'' (ない) instead of ``\textit{ar-anai}.'' 
This is implemented by \kamiya.}
In certain cases, we modified parts of the inflected forms for additional inflection beyond what this tool provides (see Table C). 
Irregular verbs were generated manually to ensure accuracy.

\input{figures/google_search_hits}
\footnotetext{To ensure visibility for forms with zero hits, we apply a smoothing technique by adding 0.5 for such cases.}

\subsection{Filtering}\label{subsec:filtering}

To ensure the correctness and actual usage of the generated inflected forms, we used SerpAPI\footnote{\url{https://serpapi.com/}} to obtain the number of exact match hits from Google search results (Table~E in Figure~\ref{fig:generate_flow}). 
Figure~\ref{fig:google_search_hits} shows the relationship between the frequency rank of inflected forms and their corresponding number of Google search hits, highlighting a long-tail distribution pattern.
We see that the trend distinctly shifts when the number of hits reaches 10.
After manually reviewing inflected forms with less than or equal to 10 hits, we concluded that most of these forms sound unnatural and should be discarded.\footnote{We release all the generated forms with their number of Google search hits for reference.}

We also manually excluded 16 specific forms that sound inappropriate with respect to honorifics.
These are respectful forms of ``\textit{shinu}'' (死ぬ, die) such as ``*\textit{o-shini-ni-naru}'' (*お死にになる) and ``*\textit{shina-reru}'' (*死なれる), which sounds inappropriate and rather unnatural. 
A more considerate expression is ``\textit{nakunar-u}'' (亡くなる, pass away), which is not registered in the current version. While there are other expressions that may not be commonly used in practice, the expressions related to ``die'' were singled out for special attention and deletion, given the need for extra caution.

%% file: figures/google_search_hits.tex
\begin{figure}
    \centering
    \includegraphics[width=7.7cm]{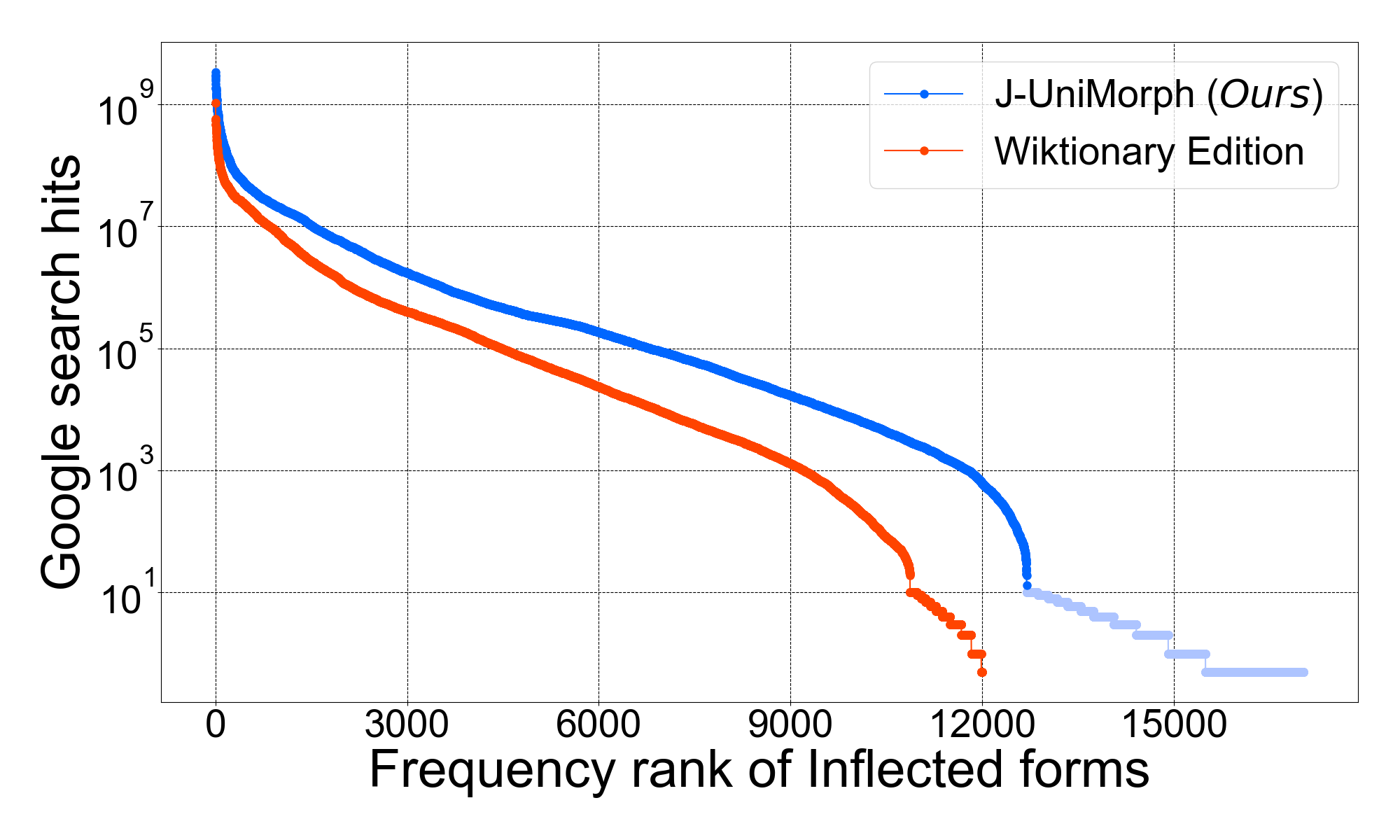}
    \caption{The relationship between the frequency rank of inflected forms and their corresponding number of Google search hits, highlighting a long-tail distribution pattern, regarding \dataname and \wik, respectively. Both graphs exhibit a clear trend shift when the number of hits falls to $10^1$ or fewer.\footnotemark~Upon manual review by authors, for \dataname, we concluded that these forms sound unnatural and should be discarded (indicated by the light-blue-colored plots), leaving a total of \total inflected forms in \dataname. Additionally, we found that inflected forms in \wik have fewer hits compared to those in \dataname (detailed in~\S\ref{subsec:4comparison}).}
    \label{fig:google_search_hits}
\end{figure}

%% file: sections/4_Comparison.tex
\section{Analysis of \dataname}
\label{sec:compare}

\subsection{Comparison with \wik}
\label{subsec:4comparison}

The SIGMORPHON–UniMorph 2023 Shared Task 0~\cite{sigmorphon:23} introduced a dataset focusing on Japanese Morphology, automatically extracted from Wiktionary.
Table~\ref{tab:wik_data} presents a list of inflection/derivation forms for the noun, ``使用'' (use-\ulabel{N}), as registered in \wik.

\input{tables/ex_wik_edition}

\input{tables/comp_with_wik_ed}

Table~\ref{tab:count_unimorph4.0japanese} shows a comparison between the \wik and~\dataname in terms of the total number of inflected forms and the number of seed words. 
\dataname has \total inflected forms in total, which slightly exceeds the number found in the \wik (12,000).
Notably, all seed words in \dataname are verbs, in contrast to \wik, where denominal verbs dominate approximately 70\%.
As explained in \S\ref{sec:feature}, inflection patterns of denominal verbs are morphologically equivalent to those of the verb ``\textit{suru}.''
It also indicates that \dataname includes a wider variety of inflection patterns and combinations, with an average of \pertotal patterns per verb, compared to the \wik, which averages 12.0.

Figure~\ref{fig:google_search_hits} presents the comparison of the number of Google search hits for all inflected forms listed in \dataname and \wik.
The graph demonstrates that \dataname contains inflected forms that are more commonly used, as indicated by higher search hits than those in \wik. The average hits by \dataname and \wik are shown in Table~\ref{tab:count_unimorph4.0japanese}.

\input{figures/visualizer}

\subsection{\dataname Visualizer}
\label{subsec:4education}

We developed the \dataname Visualizer,\footnote{\url{https://github.com/cl-tohoku/J-UniMorph}} which takes an inflected form as the input and provides the \uni labels of its form (Figure~\ref{fig:visualizer}).
This makes manual analysis of \dataname easier.
Our visualizer is different from the \kamiya by accepting input with UniMorph labels such as Past, Negative, and Polite, instead of surface forms (\textit{-ta}, \textit{-nai}, \textit{-masu}),  making it more accessible to non-native users who may not be knowledgeable about surface forms and their meanings.
We hope that this visualizer can also offer a user-friendly interface for Japanese learners, enabling them to easily understand complex Japanese verb inflection patterns.

\subsection{Labels and Forms Excluded from the Current Version}
\label{subsec:2future}

While \dataname contains a total of \total~inflected forms, covering a variety of labels and forms as described in \S{\ref{sec:feature}, we have excluded several forms, such as subsidiary verbs, question expressions, lexical causative verbs, and informal expressions.
The primary reason for their exclusion is their simple morphological pattern or morphological equivalence to other verbs already included in \dataname. 
We provide several details on the excluded forms in this section, with the detailed list available in Appendix~\ref{app:not_include}.

\paragraph{Subsidiary Verbs}

In Japanese, a small group of verbs, referred to as subsidiary verbs (\textit{hojo-d\ou shi}), are characterized by their grammaticalized functions after the \textit{te-}form.
Subsidiary verbs contribute additional meanings to the verbs they attach to.
For example, the verb ``\textit{iru}'' (いる), conveying ``be'' independently, transforms into ``be running'' or ``have run'' in the context of ``\textit{hashi-tte-iru}'' (走っている).
Similarly, the verb ``\textit{miru}'' (見る), meaning ``look'' or ``watch'' on its own, takes on a different meaning, such as ``try running,'' when attached to the verb ``\textit{hashi-ru}'' (走る, run) like ``\textit{hashi-tte-miru}'' (走ってみる). 
We generally excluded subsidiary verbs from \dataname due to their morphological equivalence to the subsidiary verbs that are already incorporated into \dataname as seed verbs.
Furthermore, one subsidiary verb can precede another subsidiary verb, to express a wide range of possible combinations, such as ``\textit{hashi-tte-mi-te-iru}'' (走ってみている). 
We set aside these patterns for future research.

\paragraph{Question Expressions}

The interrogative (\ulabel{int}) suffix ``\textit{ka}'' (か) forms questions,\footnote{In conversational contexts, raising the intonation at a sentence's end can indicate a question without a specific marker.} easily added to create inflected forms. 
However, its use with other suffixes can alter meanings. 
For example, ``\textit{tabe-masen}'' (食べません, eat-\ulabel{prs;pol;neg}), meaning ``(I) don't eat,'' becomes ``Shall (we) eat?'' when ``\textit{ka}'' is added, as in ``\textit{tabe-masen-ka}?'' (eat-\ulabel{int;inten;pol}), dropping the negation. 
Matching these combined forms with their meanings is complex, and we reserve this for future research.

\paragraph{Lexical causative verbs}

In addition to verbs that marked \ulabel{caus} (Causative) by attaching ``\textit{-se-ru/sase-ru}'' (\S\ref{subsec:2causative}), some verbs have the corresponding transitive forms that inherently carry both the causation process and the resulting event~\cite{PassCaus:11}.
Below, example (a) shows the base form ``\textit{ne-ru}'' (寝る, sleep) with the causative inflection suffix, whereas example (b) uses lexical causative verb ``\textit{nekas-u/nekas-e-ru}'' (寝かす/寝かせる) to express causative feature. 
We did not include lexical causative verbs in \dataname because they are not expressed through inflection.

\begin{itemize}
    \item[(a)] お母さんは子供を寝させた。\\
    \textit{Ok\aa san-wa, kodomo-o ne-\textbf{sase-ta}.} (``\textit{-sase-ru}'' form)\\
    The mother put the child to sleep.
    \item[(b)] お母さんは子供を寝かした/寝かせた。\\
    \textit{Ok\aa san-wa, kodomo-o \textbf{nekash-i-ta}/\textbf{nekas-e-ta}.} (lexical causative verb)\\
    The mother put the child to sleep.
\end{itemize}

\input{tables/ex_informal_form}
\paragraph{Controversial Informal Language Form}

Several colloquial expressions are controversial and seen as incorrect in Japanese.\footnote
{\url{https://www.bunka.go.jp/kokugo\_nihongo/sisaku/joho/joho/kakuki/20/tosin03/09.html}}
Table~\ref{tab:list_of_informal_form} shows examples of
omitting ``\textit{ra},'' omitting ``\textit{i},'' and inserting ``\textit{sa}.''
Although these expressions are widely used in spoken language, they are not currently used in newspapers and formal writings, and are still considered incorrect in standard language. 
Therefore, we have excluded them from the current version of \dataname.

\subsection{\uni Limitations for Japanese}
\label{subsec:4proposal}

While the UniMorph schema includes a variety of morpho-semantic features, we have identified certain Japanese expressions that are not covered by the current UniMorph labels and format.
In particular, due to its agglutinative nature, Japanese language includes compound suffixes consisting of multiple suffixes merging to express a new meaning beyond a simple combination of their individual semantic features~\cite{bunkei:89}.
For example, ``\textit{-kamo-shire-nai}'' (-かもしれない, $\approx$ maybe) consists of ``\textit{kamo}'' + ``\textit{shire}'' + ``\textit{nai}.'' 
The full meaning emerges when these suffixes are combined, with the meaning of ``\textit{nai}'' (\ulabel{neg}) disappearing in the process.

Importantly, the order of these suffixes matters.
Below, two examples showcase the same labels (\ulabel{pst}, \ulabel{pfv}, and \ulabel{lkly}) but in a different sequence.

\begin{itemize}
    \item[(a)] 彼はリンゴを食べ\textbf{たかもしれない}。\\
    \textit{Kare-wa ringo-o tabe-\textbf{ta-kamo-shire-nai}.}\\
    $\approx$ He might have eaten an apple.
    \item[(b)] 彼はリンゴを食べ\textbf{るかもしれなかった}。\\
    \textit{Kare-wa ringo-o tabe-\textbf{ru-kamo-shire-naka-tta}.}\\
    $\approx$ He could have been able to eat an apple.
\end{itemize}
In the example (a), the suffix ``\textit{-(t)ta}'' indicates \ulabel{pst;pfv} and ``\textit{-kamo-shire-}[\textit{nai|naka}]}'' represents likelihood (\ulabel{lkly}).
Although both examples contain the same set of suffixes, the meaning of each sentence differs due to the varying order of the suffixes.
That is, in example (a), \ulabel{LKLY} dominates the overall meaning more than \ulabel{PST+PFV}, whereas in example (b), \ulabel{PST+PFV} governs the overall meaning more than \ulabel{LKLY}.

One approach to address this morphological complexity is to adopt a hierarchical structure for annotations, as proposed by~\citet{guriel-etal-2022-morphological}, who explored complex argument marking in the Georgian language.

%% file: tables/ex_wik_edition.tex
\begin{table}[b!]
    \centering
    \setlength{\tabcolsep}{4pt}
    \footnotesize
    \begin{tabular}{lll}
        \toprule
        Lemma & Infleced form & Labels\\
        \midrule
        使用 & 使用せず/-\textit{se-zu} & \ulabel{V.CVB;NEG} \\
        使用 & 使用すれ/-\textit{su-re} & \ulabel{V.PTCP;IRR} \\
        使用 & 使用し/-\textit{shi} & \ulabel{V.PTCP;LGSPEC01} \\
        使用 & 使用し/-\textit{shi} & \ulabel{V.PTCP;LGSPEC02} \\
        使用 & 使用する/-\textit{suru} & \ulabel{V.PTCP;REAL} \\
        使用 & 使用すれば/-\textit{sure-ba} & *\ulabel{V;COND} \\
        使用 & 使用しよう/-\textit{shi-y\ou} & \ulabel{V;IMP;NOM(1;PL)} \\
        使用 & 使用する/-\textit{suru} & \ulabel{V;IND} \\
        使用 & 使用します/-\textit{shi-masu} & \ulabel{V;IND;FORM} \\
        使用 & 使用しない/-\textit{shi-nai} & \ulabel{V;IND;NEG} \\
        使用 & 使用される/-\textit{sa-reru} & \ulabel{V;IND;PASS} \\
        使用 & 使用した/-\textit{shi-ta} & \ulabel{V;IND;PST} \\
        使用 & 使用できる/-\textit{deki-ru} & \ulabel{V;POT} \\
        \bottomrule
    \end{tabular}
    \caption{An example of the inflection/derivation pattern for ``\textit{shiy\ou}'' (使用, use-\ulabel{N}), sourced from \wik. *使用すれば/-\textit{sure-ba} is adverb.}
    \label{tab:wik_data}
\end{table}

%Part of Wiktionary Edition
% 使用	V.CVB;NEG	使用せず
% 使用	V.PTCP;IRR	使用すれ
% 使用	V.PTCP;LGSPEC01	使用し
% 使用	V.PTCP;LGSPEC02	使用し
% 使用	V.PTCP;REAL	使用する
% 使用	V;COND	使用すれば
% 使用	V;IMP;NOM(1;PL)	使用しよう
% 使用	V;IND	使用する
% 使用	V;IND;FORM	使用します
% 使用	V;IND;NEG	使用しない
% 使用	V;IND;PASS	使用される
% 使用	V;IND;PST	使用した
% 使用	V;POT	使用できる

%% file: tables/comp_with_wik_ed.tex
\begin{table*}[t]
  \centering
  \footnotesize
  \begin{tabular}{lrrrr}
    \toprule
    & \multicolumn{3}{c}{Wiktionary Edition} & \dataname\\
    & \textit{Train} & \textit{Dev} & \textit{Test} & \textit{(Ours)}\\
    \midrule
    Number of inflected forms & 10,000 & 1,000 & 1,000 & \total\\
    \quad \textbf{Number of inflected forms per word} & \textbf{12.5} & \textbf{10.0} & \textbf{10.0} & \textbf{\pertotal}\\
    \quad The average of number of hits (in millions) & 3.4 & 4.6 & 5.5 & 12\\
    \midrule
    Number of seed words & 800 & 100 & 100 & 107\\
    \quad \textbf{Verbs} & \textbf{25\%} & \textbf{27\%} & \textbf{30\%} & \textbf{100\%}\\
    \quad Denominal verbs (noun + ``\textit{suru}'') & 72\% & 69\% & 67\% & 0\%\\
    \quad Accompanied by particles & 3\% & 2\% & 3\% & 0\%\\
    \quad Deadverbal verbs (adverb + ``\textit{suru}'') & 1\% & 2\% & 0\% & 0\%\\
    \bottomrule
  \end{tabular}
  \caption{Comparison of lemma types between Wiktionary Edition and \dataname.}
  \label{tab:count_unimorph4.0japanese}
\end{table*}

%% file: figures/visualizer.tex
\begin{figure}[t]
    \centering
    \includegraphics[width=7.7cm]{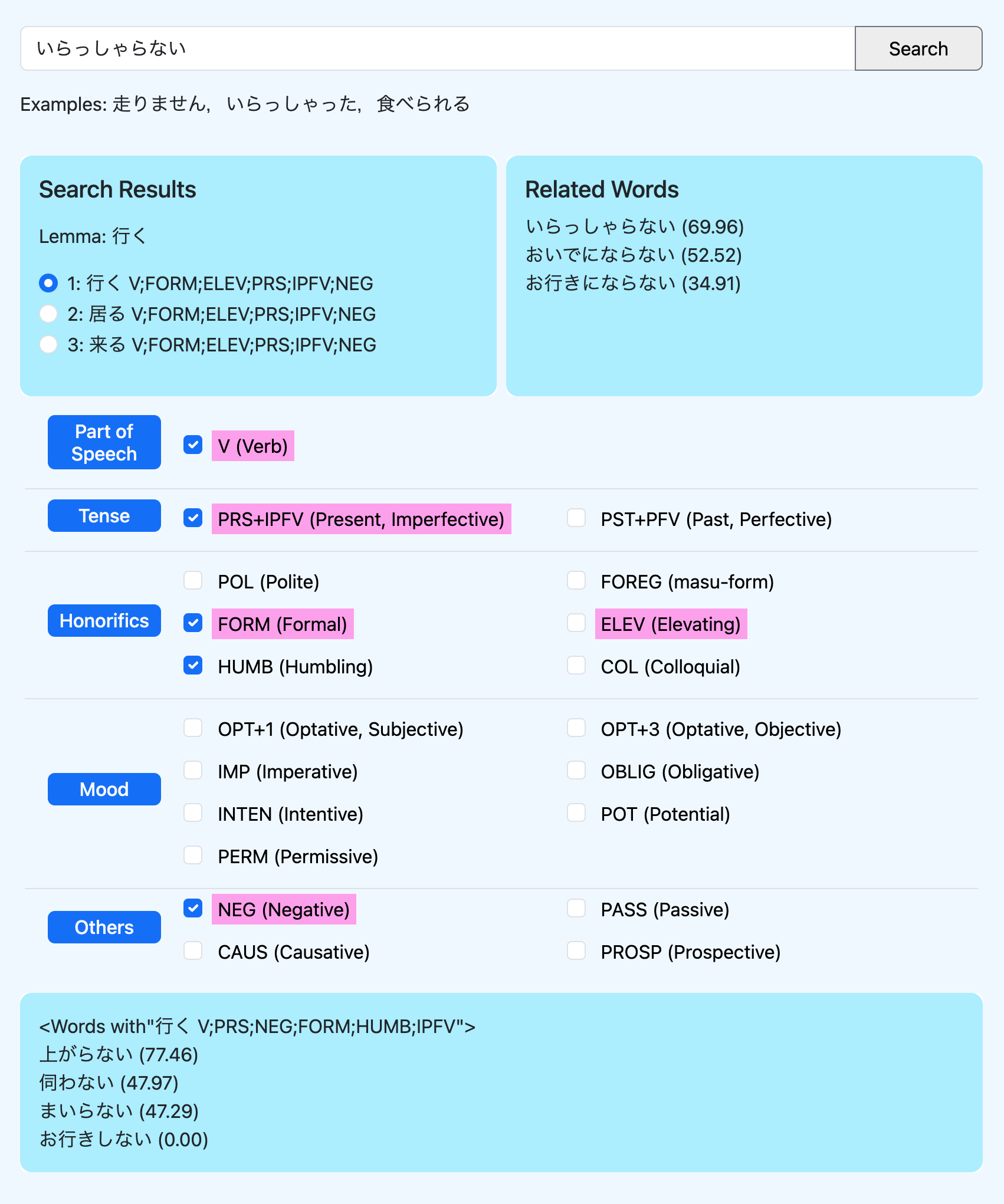}
    \caption{Screenshot of \dataname Visualizer, a tool for helping Japanese learners. Users input an inflected form and click the ``Search'' button to highlight corresponding \uni labels. If the inflected form has multiple meanings, they are displayed under the ``Search Results'' section, with the option to toggle between meanings. Additionally, ``Related Words'' section displays other inflected forms with the same label (including itself). Confidence values, ranging from 0 to 100 and based on Google search hits, assist users in determining which inflected form should be used. Higher values indicate more hits. Users also can switch between labels to investigate inflected forms with different meanings.}
    \label{fig:visualizer}
\end{figure}

%% file: tables/ex_informal_form.tex
\begin{table*}[t]
  \centering
  \resizebox{\textwidth}{!}{
  \begin{tabular}{llll}
    \toprule
    Category & Formal Form & Informal Form & Rough translation \\
    \midrule
    Omitting \textit{ra} & \textit{tabe-\textbf{ra}reru}/食べ\textbf{ら}れる & \textit{tabe-reru}/食べれる & can eat \\
    Omitting \textit{i} & \textit{tabe-te-\textbf{i}ru}/食べて\textbf{い}る & \textit{tabe-te-ru}/食べてる & be eating, have eaten \\
    Inserting \textit{sa} & \textit{kawa-sete-itadaku}/買わせていただく & \textit{kawa-\textbf{sa}-sete-itadaku}/買わ\textbf{さ}せていただく & have the honor of buying \\
    \bottomrule
  \end{tabular}
  }
  \caption{Examples of Informal Forms}
  \label{tab:list_of_informal_form}
\end{table*}

%% file: sections/5_Conclusion.tex
\section{Conclusion}\label{sec:concl}

We introduced \dataname, a Japanese Morphology dataset based on the \uni schema.
\dataname covers a wide range of verb inflection forms, including honorifics, politeness levels, and other linguistic nuances, reflecting the language's agglutinative nature. 
Unlike the Wiktionary Edition, which is automatically extracted from Wiktionary, \dataname has been carefully designed by native speakers, featuring an average of 118 inflected forms per word (with a total of \total instances), compared to Wiktionary Edition's 12 inflected forms per word (12,000 instances in total). 
\dataname, along with its interactive visualizer, has been released to facilitate cross-linguistic research and applications, offering a more comprehensive resource than previously available.

%% file: sections/acknowledgments.tex
\section*{Acknowledgments}

This work was supported by RIKEN Special Postdoctoral Researchers Program and JSPS KAKENHI Grant Numbers JP21K21343, JP22H00524.

While conducting this research, we received valuable comments from members of the Tohoku NLP Group at Tohoku University, Japan. We deeply appreciate their support and insightful contributions.

%% file: sections/appendix.tex
\onecolumn

\section{Correspondence between the basic form and the lexical honorifics}
\label{app:keigo}

%Figure \ref{fig:keigo_corres} shows the correspondence between the basic forms and the lexical honorifics adopted in \dataname.

\input{figures/keigo_corres}

\section{Statistics of generated inflected forms in \dataname}
\label{app:verb_count}

% Table \ref{tab:count_surface_forms} summarizes the number of verbs and generated inflected forms per verb for each conjugation type.

\input{tables/appB_count_surface_forms}

\clearpage

\section{Inflection/derivation affixes not included in \dataname}
\label{app:not_include}

% Table \ref{tab:list_of_not_generate} lists the future work that we did not adopt in \dataname.
% Each of these forms has past, negative, polite, honorific, and other expressions, and some of these are unnatural.

\input{tables/appC_list_of_not_generate}

\twocolumn

%% file: figures/keigo_corres.tex
\begin{figure*}[h]
    \centering
    \includegraphics[width=16cm]{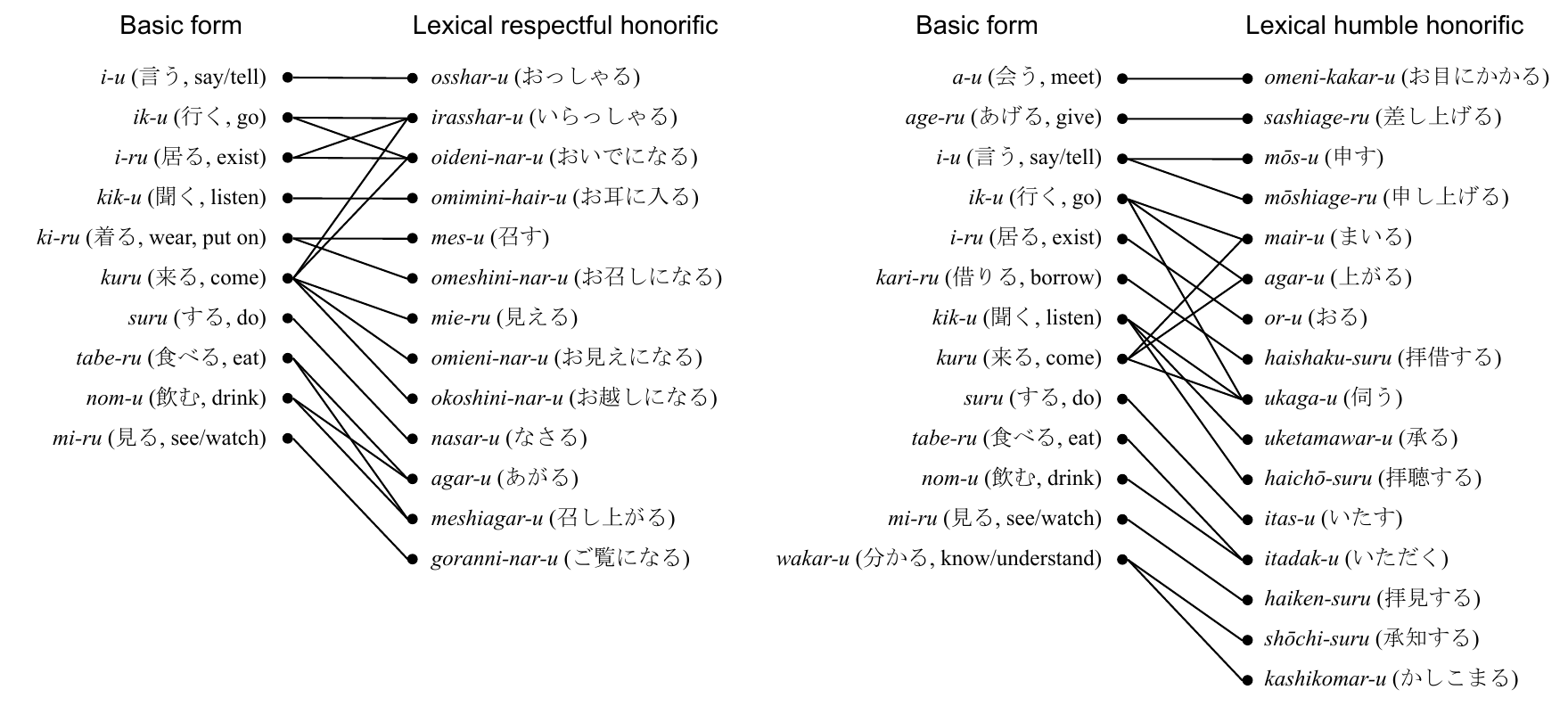}
    \caption{Correspondence between the basic forms and the lexical honorifics adopted in \dataname.}
    \label{fig:keigo_corres}
\end{figure*}

%% file: tables/appB_count_surface_forms.tex
\begin{table*}[h]
  \centering
  \footnotesize
  \begin{tabular}{llrr}
    \toprule
Politeness Type & Conjugation Type & Verbs & Generated inflected forms \\
\midrule
Basic & Regular I & 76 & 126 \\
 & Regular II & 29 & 118 \\
 & ``\textit{kuru}'' (Irregular) & 1 & 100 \\
 & ``\textit{suru}'' (Irregular) & 1 & 102 \\
\midrule
Lexical respectful honorifics & Regular I & 18 & 103 \\
 & Regular II & 1 & 94 \\
\midrule
Lexical humble honorifics & Regular I & 15 & 92 \\
 & Regular II & 2 & 84 \\
 & ``\textit{-suru}'' (Irregular) & 4 & 84 \\
\bottomrule
  \end{tabular}
  \caption{The number of verbs and generated inflected forms per verb for each conjugation type. The numbers represent the counts prior to excluding infrequent inflected forms.}
  \label{tab:count_surface_forms}
\end{table*}

%% file: tables/appC_list_of_not_generate.tex
\begin{table*}[ht]
  \centering
  \footnotesize
  \setlength{\tabcolsep}{5pt}
  \begin{tabular}{lll}
  \toprule
Reason & Affixes or example Inflected forms & Romanized and Rough translation\\
  \midrule
Subsidiary verbs (補助動詞) & 〜ている & \textit{-te-iru} (be doing, have done)\\
 & 〜てみる & \textit{-te-miru} (try doing)\\
 & 〜ておく & \textit{-te-oku} (do in advance)\\
 & 〜ておこう & \textit{-te-ok\ou} (let's do in advance)\\
 & 〜てあげる & \textit{-te-ageru} (do something for the benefit of someone)\\
 & 〜てもらう & \textit{-te-morau} (get someone to do something)\\
 & 〜てくれる & \textit{-te-kureru} (someone do something for me/us)\\
 & 〜てある & \textit{-te-aru} (has been done)\\
 & 〜てしまう & \textit{-te-shimau} (end up doing)\\
 & 〜ていく & \textit{-te-iku} (keep on doing)\\
 & 〜つつある & \textit{-tsutsu-aru} (be about to do)\\
 & 〜てほしい & \textit{-te-hosh\ii} (want someone to do)\\
\midrule
Compound suffixes (複合辞) & 〜かもしれない & \textit{-kamo-shire-nai} (may)\\
 & 〜てはいけない & \textit{-tewa-ike-nai} (must not do)\\
 & 〜てはならない & \textit{-tewa-nara-nai} (must not do)\\
 & 〜たがっている & \textit{-tagatte-iru} (wants to do)\\
 & 〜なければならない & \textit{-nakereba-naranai} (have to do)\\
 & 〜に違いない & \textit{-ni-chigai-nai} (must be doing)\\
\midrule
Non verbs & 〜てもいい & \textit{-te-mo-\ii} (permissive)\\
 & 〜たら & \textit{-tara} (if)\\
 & 〜たり & \textit{-tari} (do and ...)\\
 & 〜ば & \textit{-ba} (if) \\
 & 〜べきだ & \textit{-beki-da} (should)\\
 & 〜つもりだ & \textit{-tsumori-da} (intend to do)\\
 & 〜はずだ & \textit{-hazu-da} (be supposed to do)\\
 & 〜らしい & \textit{-rash\ii} (It seems like ...)\\
 & 〜べし & \textit{-beshi} (should do)\\
 & 〜べからず & \textit{-bekara-zu} (should not do)\\
 & 「笑い」「話」など & \begin{tabular}{l} Treat as nouns, such as \\ \textit{warai} (laughter), \textit{hanashi} (talk/conversation) \end{tabular} \\
 & 〜に〜（「買いに行く」など） & \textit{-ni-} (adverbial usage) \\
 & 〜ながら & \textit{-nagara} (while doing)\\
 & 〜そうだ & \textit{-s\ou da} (It seems like ...)\\
 & 〜物，〜方 & \textit{-mono}, \textit{-kata} (Nominative usage)\\
 & 〜始める，〜終わる & \textit{-hajimeru}, \textit{-owaru} (begin -ing, finish -ing)\\
\midrule
Noun/Adverb + light verb & 〜する & \textit{-suru} (light verb) \\
\midrule
Lexical causative verbs & 寝かせる，立てる & \textit{nekaseru}, \textit{tateru}\\
\midrule
Omitting \textit{ra} (ら抜き言葉) & 〜れる & \textit{-reru} \\
Omitting \textit{i} (い抜き言葉 & 〜てる & \textit{-teru} \\
Inserting \textit{sa} (さ入れ言葉) & 〜させて〜 & \textit{-sase-te-} \\
\midrule
Interrogative suffix & 〜か？ & \textit{-ka?} \\
 & 〜ましょうか？，〜ませんか？ & \textit{-mash\ou ka?}, \textit{masen-ka?}\\
\midrule
Another respectful expressions & お〜くださる & \textit{o---kudasaru}\\
 & お〜なさる & \textit{o---nasaru} \\
Another humble expressions & お〜いたす & \textit{o---itasu}\\
 & お〜いたします & \textit{o---itashi-masu} \\
\midrule
Others & 〜れる/られる & \textit{-reru/rareru} (spontaneous) \\
 & 〜よう & \textit{-y\ou} (speculation) \\
  \bottomrule
  \end{tabular}
  \caption{List of inflection/derivation affixes not included in the current version of \dataname.}
  \label{tab:list_of_not_generate}
  \vspace{-25mm}
\end{table*}

%% file: main.bbl
\begin{thebibliography}{21}
\expandafter\ifx\csname natexlab\endcsname\relax\def\natexlab#1{#1}\fi

\bibitem[{Batsuren et~al.(2022)Batsuren, Goldman, Khalifa, Habash, Kiera{\'s}, Bella, Leonard, Nicolai, Gorman, Ate, Ryskina, Mielke, Budianskaya, El-Khaissi, Pimentel, Gasser, Lane, Raj, Coler, Samame, Camaiteri, Rojas, L{\'o}pez~Francis, Oncevay, L{\'o}pez~Bautista, Villegas, Hennigen, Ek, Guriel, Dirix, Bernardy, Scherbakov, Bayyr-ool, Anastasopoulos, Zariquiey, Sheifer, Ganieva, Cruz, Karah{\'o}{\v{g}}a, Markantonatou, Pavlidis, Plugaryov, Klyachko, Salehi, Angulo, Baxi, Krizhanovsky, Krizhanovskaya, Salesky, Vania, Ivanova, White, Maudslay, Valvoda, Zmigrod, Czarnowska, Nikkarinen, Salchak, Bhatt, Straughn, Liu, Washington, Pinter, Ataman, Wolinski, Suhardijanto, Yablonskaya, Stoehr, Dolatian, Nuriah, Ratan, Tyers, Ponti, Aiton, Arora, Hatcher, Kumar, Young, Rodionova, Yemelina, Andrushko, Marchenko, Mashkovtseva, Serova, Prud{'}hommeaux, Nepomniashchaya, Giunchiglia, Chodroff, Hulden, Silfverberg, McCarthy, Yarowsky, Cotterell, Tsarfaty, and Vylomova}]{UniMorph4.0:22}
Khuyagbaatar Batsuren, Omer Goldman, Salam Khalifa, Nizar Habash, Witold Kiera{\'s}, G{\'a}bor Bella, Brian Leonard, Garrett Nicolai, Kyle Gorman, Yustinus~Ghanggo Ate, Maria Ryskina, Sabrina Mielke, Elena Budianskaya, Charbel El-Khaissi, Tiago Pimentel, Michael Gasser, William~Abbott Lane, Mohit Raj, Matt Coler, Jaime Rafael~Montoya Samame, Delio~Siticonatzi Camaiteri, Esa{\'u}~Zumaeta Rojas, Didier L{\'o}pez~Francis, Arturo Oncevay, Juan L{\'o}pez~Bautista, Gema Celeste~Silva Villegas, Lucas~Torroba Hennigen, Adam Ek, David Guriel, Peter Dirix, Jean-Philippe Bernardy, Andrey Scherbakov, Aziyana Bayyr-ool, Antonios Anastasopoulos, Roberto Zariquiey, Karina Sheifer, Sofya Ganieva, Hilaria Cruz, Ritv{\'a}n Karah{\'o}{\v{g}}a, Stella Markantonatou, George Pavlidis, Matvey Plugaryov, Elena Klyachko, Ali Salehi, Candy Angulo, Jatayu Baxi, Andrew Krizhanovsky, Natalia Krizhanovskaya, Elizabeth Salesky, Clara Vania, Sardana Ivanova, Jennifer White, Rowan~Hall Maudslay, Josef Valvoda, Ran Zmigrod, Paula Czarnowska,
  Irene Nikkarinen, Aelita Salchak, Brijesh Bhatt, Christopher Straughn, Zoey Liu, Jonathan~North Washington, Yuval Pinter, Duygu Ataman, Marcin Wolinski, Totok Suhardijanto, Anna Yablonskaya, Niklas Stoehr, Hossep Dolatian, Zahroh Nuriah, Shyam Ratan, Francis~M. Tyers, Edoardo~M. Ponti, Grant Aiton, Aryaman Arora, Richard~J. Hatcher, Ritesh Kumar, Jeremiah Young, Daria Rodionova, Anastasia Yemelina, Taras Andrushko, Igor Marchenko, Polina Mashkovtseva, Alexandra Serova, Emily Prud{'}hommeaux, Maria Nepomniashchaya, Fausto Giunchiglia, Eleanor Chodroff, Mans Hulden, Miikka Silfverberg, Arya~D. McCarthy, David Yarowsky, Ryan Cotterell, Reut Tsarfaty, and Ekaterina Vylomova. 2022.
\newblock \href {https://aclanthology.org/2022.lrec-1.89} {{U}ni{M}orph 4.0: {U}niversal {M}orphology}.
\newblock In \emph{Proceedings of the Thirteenth Language Resources and Evaluation Conference}, pages 840--855, Marseille, France. European Language Resources Association.

\bibitem[{Cotterell et~al.(2018)Cotterell, Kirov, Sylak-Glassman, Walther, Vylomova, McCarthy, Kann, Mielke, Nicolai, Silfverberg, Yarowsky, Eisner, and Hulden}]{sigmorphon:18}
Ryan Cotterell, Christo Kirov, John Sylak-Glassman, G{\'e}raldine Walther, Ekaterina Vylomova, Arya~D. McCarthy, Katharina Kann, Sabrina~J. Mielke, Garrett Nicolai, Miikka Silfverberg, David Yarowsky, Jason Eisner, and Mans Hulden. 2018.
\newblock \href {https://doi.org/10.18653/v1/K18-3001} {The {C}o{NLL}{--}{SIGMORPHON} 2018 shared task: Universal morphological reinflection}.
\newblock In \emph{Proceedings of the {C}o{NLL}{--}{SIGMORPHON} 2018 Shared Task: Universal Morphological Reinflection}, pages 1--27, Brussels. Association for Computational Linguistics.

\bibitem[{Cotterell et~al.(2017)Cotterell, Kirov, Sylak-Glassman, Walther, Vylomova, Xia, Faruqui, K{\"u}bler, Yarowsky, Eisner, and Hulden}]{sigmorphon:17}
Ryan Cotterell, Christo Kirov, John Sylak-Glassman, G{\'e}raldine Walther, Ekaterina Vylomova, Patrick Xia, Manaal Faruqui, Sandra K{\"u}bler, David Yarowsky, Jason Eisner, and Mans Hulden. 2017.
\newblock \href {https://doi.org/10.18653/v1/K17-2001} {{C}o{NLL}-{SIGMORPHON} 2017 shared task: Universal morphological reinflection in 52 languages}.
\newblock In \emph{Proceedings of the {C}o{NLL} {SIGMORPHON} 2017 Shared Task: Universal Morphological Reinflection}, pages 1--30, Vancouver. Association for Computational Linguistics.

\bibitem[{Cotterell et~al.(2016)Cotterell, Kirov, Sylak-Glassman, Yarowsky, Eisner, and Hulden}]{sigmorphon:16}
Ryan Cotterell, Christo Kirov, John Sylak-Glassman, David Yarowsky, Jason Eisner, and Mans Hulden. 2016.
\newblock \href {https://doi.org/10.18653/v1/W16-2002} {The {SIGMORPHON} 2016 shared {T}ask{---}{M}orphological reinflection}.
\newblock In \emph{Proceedings of the 14th {SIGMORPHON} Workshop on Computational Research in Phonetics, Phonology, and Morphology}, pages 10--22, Berlin, Germany. Association for Computational Linguistics.

\bibitem[{Goldman et~al.(2023)Goldman, Batsuren, Khalifa, Arora, Nicolai, Tsarfaty, and Vylomova}]{sigmorphon:23}
Omer Goldman, Khuyagbaatar Batsuren, Salam Khalifa, Aryaman Arora, Garrett Nicolai, Reut Tsarfaty, and Ekaterina Vylomova. 2023.
\newblock \href {https://doi.org/10.18653/v1/2023.sigmorphon-1.13} {{SIGMORPHON}{--}{U}ni{M}orph 2023 shared task 0: Typologically diverse morphological inflection}.
\newblock In \emph{Proceedings of the 20th SIGMORPHON workshop on Computational Research in Phonetics, Phonology, and Morphology}, pages 117--125, Toronto, Canada. Association for Computational Linguistics.

\bibitem[{Goldman and Tsarfaty(2022)}]{MightyMorph:22}
Omer Goldman and Reut Tsarfaty. 2022.
\newblock \href {https://doi.org/10.1162/tacl_a_00528} {{Morphology Without Borders: Clause-Level Morphology}}.
\newblock \emph{Transactions of the Association for Computational Linguistics}, 10:1455--1472.

\bibitem[{Guriel et~al.(2022)Guriel, Goldman, and Tsarfaty}]{guriel-etal-2022-morphological}
David Guriel, Omer Goldman, and Reut Tsarfaty. 2022.
\newblock \href {https://aclanthology.org/2022.acl-short.21} {Morphological reinflection with multiple arguments: An extended annotation schema and a {G}eorgian case study}.
\newblock In \emph{Proceedings of the 60th Annual Meeting of the Association for Computational Linguistics (Volume 2: Short Papers)}, pages 196--202, Dublin, Ireland. Association for Computational Linguistics.

\bibitem[{Hirabayashi and Hama(1988)}]{keigo:88}
Yoshisuke Hirabayashi and Yumiko Hama. 1988.
\newblock \emph{{Keigo (Honorific Speech)}}.
\newblock Aratake Publishers.

\bibitem[{Kamiya(2001)}]{KamiyaVerb:01}
Taeko Kamiya. 2001.
\newblock \emph{The handbook of Japanese verbs}.
\newblock Kodansha.

\bibitem[{Kato and Fukuchi(1989)}]{tense_aspect:89}
Yasuhiko Kato and Tsutomu Fukuchi. 1989.
\newblock \emph{Tense, Aspect, and Mood}.
\newblock Aratake Publishers.

\bibitem[{Kodner et~al.(2022)Kodner, Khalifa, Batsuren, Dolatian, Cotterell, Akkus, Anastasopoulos, Andrushko, Arora, Atanalov, Bella, Budianskaya, Ghanggo~Ate, Goldman, Guriel, Guriel, Guriel-Agiashvili, Kiera{\'s}, Krizhanovsky, Krizhanovsky, Marchenko, Markowska, Mashkovtseva, Nepomniashchaya, Rodionova, Scheifer, Sorova, Yemelina, Young, and Vylomova}]{sigmorphon:22}
Jordan Kodner, Salam Khalifa, Khuyagbaatar Batsuren, Hossep Dolatian, Ryan Cotterell, Faruk Akkus, Antonios Anastasopoulos, Taras Andrushko, Aryaman Arora, Nona Atanalov, G{\'a}bor Bella, Elena Budianskaya, Yustinus Ghanggo~Ate, Omer Goldman, David Guriel, Simon Guriel, Silvia Guriel-Agiashvili, Witold Kiera{\'s}, Andrew Krizhanovsky, Natalia Krizhanovsky, Igor Marchenko, Magdalena Markowska, Polina Mashkovtseva, Maria Nepomniashchaya, Daria Rodionova, Karina Scheifer, Alexandra Sorova, Anastasia Yemelina, Jeremiah Young, and Ekaterina Vylomova. 2022.
\newblock \href {https://doi.org/10.18653/v1/2022.sigmorphon-1.19} {{SIGMORPHON}{--}{U}ni{M}orph 2022 shared task 0: Generalization and typologically diverse morphological inflection}.
\newblock In \emph{Proceedings of the 19th SIGMORPHON Workshop on Computational Research in Phonetics, Phonology, and Morphology}, pages 176--203, Seattle, Washington. Association for Computational Linguistics.

\bibitem[{McCarthy et~al.(2020)McCarthy, Kirov, Grella, Nidhi, Xia, Gorman, Vylomova, Mielke, Nicolai, Silfverberg, Arkhangelskiy, Krizhanovsky, Krizhanovsky, Klyachko, Sorokin, Mansfield, Ern{\v{s}}treits, Pinter, Jacobs, Cotterell, Hulden, and Yarowsky}]{UniMorph3.0:20}
Arya~D. McCarthy, Christo Kirov, Matteo Grella, Amrit Nidhi, Patrick Xia, Kyle Gorman, Ekaterina Vylomova, Sabrina~J. Mielke, Garrett Nicolai, Miikka Silfverberg, Timofey Arkhangelskiy, Nataly Krizhanovsky, Andrew Krizhanovsky, Elena Klyachko, Alexey Sorokin, John Mansfield, Valts Ern{\v{s}}treits, Yuval Pinter, Cassandra~L. Jacobs, Ryan Cotterell, Mans Hulden, and David Yarowsky. 2020.
\newblock \href {https://aclanthology.org/2020.lrec-1.483} {{U}ni{M}orph 3.0: {U}niversal {M}orphology}.
\newblock In \emph{Proceedings of the Twelfth Language Resources and Evaluation Conference}, pages 3922--3931, Marseille, France. European Language Resources Association.

\bibitem[{McCarthy et~al.(2019)McCarthy, Vylomova, Wu, Malaviya, Wolf-Sonkin, Nicolai, Kirov, Silfverberg, Mielke, Heinz, Cotterell, and Hulden}]{sigmorphon:19}
Arya~D. McCarthy, Ekaterina Vylomova, Shijie Wu, Chaitanya Malaviya, Lawrence Wolf-Sonkin, Garrett Nicolai, Christo Kirov, Miikka Silfverberg, Sabrina~J. Mielke, Jeffrey Heinz, Ryan Cotterell, and Mans Hulden. 2019.
\newblock \href {https://doi.org/10.18653/v1/W19-4226} {The {SIGMORPHON} 2019 shared task: Morphological analysis in context and cross-lingual transfer for inflection}.
\newblock In \emph{Proceedings of the 16th Workshop on Computational Research in Phonetics, Phonology, and Morphology}, pages 229--244, Florence, Italy. Association for Computational Linguistics.

\bibitem[{Morita and Matsuki(1989)}]{bunkei:89}
Yoshiyuki Morita and Masae Matsuki. 1989.
\newblock \emph{{Nihongo Hyogen Bunkei (Structures of Japanese Expressions)}}.
\newblock ALC PRESS.

\bibitem[{{Nihongo Kijutsu Bunpo Kenkyukai}(2007)}]{Nitta3:07}
{Nihongo Kijutsu Bunpo Kenkyukai}. 2007.
\newblock \emph{{Gendai Nihongo Bunpo 3 (Contemporary Japanese Grammar 3)}}.
\newblock Kurosio Publishers.
\newblock (In Japanese).

\bibitem[{{Nihongo Kijutsu Bunpo Kenkyukai}(2009{\natexlab{a}})}]{Nitta2:09}
{Nihongo Kijutsu Bunpo Kenkyukai}. 2009{\natexlab{a}}.
\newblock \emph{{Gendai Nihongo Bunpo 2 (Contemporary Japanese Grammar 2)}}.
\newblock Kurosio Publishers.
\newblock (In Japanese).

\bibitem[{{Nihongo Kijutsu Bunpo Kenkyukai}(2009{\natexlab{b}})}]{Nitta7:09}
{Nihongo Kijutsu Bunpo Kenkyukai}. 2009{\natexlab{b}}.
\newblock \emph{{Gendai Nihongo Bunpo 7 (Contemporary Japanese Grammar 7)}}.
\newblock Kurosio Publishers.
\newblock (In Japanese).

\bibitem[{Pimentel et~al.(2021)Pimentel, Ryskina, Mielke, Wu, Chodroff, Leonard, Nicolai, Ghanggo~Ate, Khalifa, Habash, El-Khaissi, Goldman, Gasser, Lane, Coler, Oncevay, Montoya~Samame, Silva~Villegas, Ek, Bernardy, Shcherbakov, Bayyr-ool, Sheifer, Ganieva, Plugaryov, Klyachko, Salehi, Krizhanovsky, Krizhanovsky, Vania, Ivanova, Salchak, Straughn, Liu, Washington, Ataman, Kiera{\'s}, Woli{\'n}ski, Suhardijanto, Stoehr, Nuriah, Ratan, Tyers, Ponti, Aiton, Hatcher, Prud{'}hommeaux, Kumar, Hulden, Barta, Lakatos, Szolnok, {\'A}cs, Raj, Yarowsky, Cotterell, Ambridge, and Vylomova}]{sigmorphon:21}
Tiago Pimentel, Maria Ryskina, Sabrina~J. Mielke, Shijie Wu, Eleanor Chodroff, Brian Leonard, Garrett Nicolai, Yustinus Ghanggo~Ate, Salam Khalifa, Nizar Habash, Charbel El-Khaissi, Omer Goldman, Michael Gasser, William Lane, Matt Coler, Arturo Oncevay, Jaime~Rafael Montoya~Samame, Gema~Celeste Silva~Villegas, Adam Ek, Jean-Philippe Bernardy, Andrey Shcherbakov, Aziyana Bayyr-ool, Karina Sheifer, Sofya Ganieva, Matvey Plugaryov, Elena Klyachko, Ali Salehi, Andrew Krizhanovsky, Natalia Krizhanovsky, Clara Vania, Sardana Ivanova, Aelita Salchak, Christopher Straughn, Zoey Liu, Jonathan~North Washington, Duygu Ataman, Witold Kiera{\'s}, Marcin Woli{\'n}ski, Totok Suhardijanto, Niklas Stoehr, Zahroh Nuriah, Shyam Ratan, Francis~M. Tyers, Edoardo~M. Ponti, Grant Aiton, Richard~J. Hatcher, Emily Prud{'}hommeaux, Ritesh Kumar, Mans Hulden, Botond Barta, Dorina Lakatos, G{\'a}bor Szolnok, Judit {\'A}cs, Mohit Raj, David Yarowsky, Ryan Cotterell, Ben Ambridge, and Ekaterina Vylomova. 2021.
\newblock \href {https://doi.org/10.18653/v1/2021.sigmorphon-1.25} {{SIGMORPHON} 2021 shared task on morphological reinflection: Generalization across languages}.
\newblock In \emph{Proceedings of the 18th SIGMORPHON Workshop on Computational Research in Phonetics, Phonology, and Morphology}, pages 229--259, Online. Association for Computational Linguistics.

\bibitem[{Sylak-Glassman(2016)}]{schema:16}
John Sylak-Glassman. 2016.
\newblock The composition and use of the universal morphological feature schema (unimorph schema).
\newblock \emph{Johns Hopkins University}.

\bibitem[{Takami(2011)}]{PassCaus:11}
Ken-ichi Takami. 2011.
\newblock \emph{Ukemi to Shieki (Passive and Causative)}.
\newblock Kaitakusha.

\bibitem[{Vylomova et~al.(2020)Vylomova, White, Salesky, Mielke, Wu, Ponti, Maudslay, Zmigrod, Valvoda, Toldova, Tyers, Klyachko, Yegorov, Krizhanovsky, Czarnowska, Nikkarinen, Krizhanovsky, Pimentel, Torroba~Hennigen, Kirov, Nicolai, Williams, Anastasopoulos, Cruz, Chodroff, Cotterell, Silfverberg, and Hulden}]{sigmorphon:20}
Ekaterina Vylomova, Jennifer White, Elizabeth Salesky, Sabrina~J. Mielke, Shijie Wu, Edoardo~Maria Ponti, Rowan~Hall Maudslay, Ran Zmigrod, Josef Valvoda, Svetlana Toldova, Francis Tyers, Elena Klyachko, Ilya Yegorov, Natalia Krizhanovsky, Paula Czarnowska, Irene Nikkarinen, Andrew Krizhanovsky, Tiago Pimentel, Lucas Torroba~Hennigen, Christo Kirov, Garrett Nicolai, Adina Williams, Antonios Anastasopoulos, Hilaria Cruz, Eleanor Chodroff, Ryan Cotterell, Miikka Silfverberg, and Mans Hulden. 2020.
\newblock \href {https://doi.org/10.18653/v1/2020.sigmorphon-1.1} {{SIGMORPHON} 2020 shared task 0: Typologically diverse morphological inflection}.
\newblock In \emph{Proceedings of the 17th SIGMORPHON Workshop on Computational Research in Phonetics, Phonology, and Morphology}, pages 1--39, Online. Association for Computational Linguistics.

\end{thebibliography}
